# Supervised Machine Learning with a Novel Kernel Density Estimator (arXiv:0709.2760)

Yen-Jen Oyang[*,1], Darby Tien-Hao Chang[2], Yu-Yen Ou[3],
Hao-Geng Hung[4], Chih-Peng Wu[4], and Chien-Yu Chen[5]

[1] Graduate Institute of Biomedical Electronics and Bioinformatics, Department of Computer Science and Information Engineering, National Taiwan University, Taipei, 106, Taiwan, R.O.C. yjoyang@csie.ntu.edu.tw
[2] Department of Electrical Engineering, National Cheng Kung University, Tainan, 70101, Taiwan, R.O.C. darby@ee.ncku.edu.tw
[3] Graduate School of Biotechnology and Bioinformatics, Yuan-Ze University, Chung-Li, 320, Taiwan, R.O.C. yien@saturn.yzu.edu.tw
[4] Department of Computer Science and Information Engineering, National Taiwan University, Taipei, 106, Taiwan, R.O.C. hghung@mars.csie.ntu.edu.tw, chinuy@gmail.com
[5] Department of Bio-Industrial Mechatronics Engineering, National Taiwan University, Taipei, Taiwan, R.O.C. chienyuchen@ntu.edu.tw

**Abstract.** In recent years, kernel density estimation has been exploited by computer scientists to model machine learning problems. The kernel density estimation based approaches are of interest due to the low time complexity of either $O(n)$ or $O(n \log n)$ for constructing a classifier, where $n$ is the number of sampling instances. Concerning design of kernel density estimators, one essential issue is how fast the pointwise mean square error (MSE) and/or the integrated mean square error (IMSE) diminish as the number of sampling instances increases. In this article, the kernel function with general form $\frac{\sqrt{2} \cdot \Gamma(m/2+1)}{\pi^{(m+1)/2}} \cdot \frac{1}{\sigma} \cdot \exp\left(\frac{-\left(x_1^2 + x_2^2 + ... + x_m^2\right)^m}{2\sigma^2}\right)$, where $m$ is the dimension of the vector space, is employed for generation of the density estimator in a high-dimensional vector space. With the proposed kernel function, it is then feasible to make the pointwise MSE of the density estimator converge at $O(n^{-2/3})$ regardless of the dimension of the vector space, provided that the probability density function at the point of interest meets certain conditions.

**Keyterms:** kernel density estimation, machine learning, data classification

---

[*] To whom correspondences should be addressed. Tel:+886-2-33664888 ext. 431, Fax:+886-2-23688675.

## I. Introduction

Kernel density estimation is a problem that has been studied by statisticians for decades [1-4]. In recent years, kernel density estimation has been exploited by computer scientists to model machine learning problems [5-7]. The kernel density estimation based approaches are of interest due to the low time complexity of either $O(n)$ or $O(n \log n)$ for generating an estimator, where *n* is the number of sampling instances [4]. Furthermore, in comparison with the support vector machine (SVM) [8], a recent study has shown that the kernel density estimation based classifier is capable of delivering the same level of prediction accuracy, while enjoying several distinctive advantages [7]. Therefore, the kernel density estimation based machine learning algorithms may become the favorite choice for contemporary applications that involve large datasets or databases.

Concerning design of kernel density estimators, one essential issue is how fast the pointwise mean square error (MSE) and/or the integrated mean square error (IMSE) diminish as the number of sampling instances increases. In this respect, the main problems with the conventional kernel density estimators is that the convergence rate of the pointwise MSE becomes extremely slow in case the dimension of the dataset is large. For example, with Gaussian kernels, the pointwise MSE of the fixed kernel density estimator converges at $O(n^{-4/(m+4)})$ [4], where *m* is the dimension of the dataset. Accordingly, the conventional kernel density estimators suffer a serious deficiency in dealing with high-dimensional datasets. Since high-dimensional datasets are common in modern machine learning applications, design of a novel kernel density estimator that can handle high-dimensional datasets more effectively is essential for exploiting kernel density estimation in modern machine learning applications.

In this article, the kernel function with general form

$$\frac{\sqrt{2} \cdot \Gamma(m/2+1)}{\pi^{(m+1)/2}} \cdot \frac{1}{\sigma} \cdot \exp\left(\frac{-\left(x_1^2 + x_2^2 + \ldots + x_m^2\right)^m}{2\sigma^2}\right),$$

where *m* is the dimension of the vector space, is employed for generation of the density estimator in a high-dimensional vector space. With the proposed kernel function, it is then feasible to make the pointwise MSE of the density estimator converge at $O(n^{-2/3})$ regardless of the dimension of the vector space, provided that the probability density function at the point of interest meets certain conditions. Just like many conventional kernel density estimators, the proposed kernel density estimator features an average time complexity of $O(n \log n)$ for generating the approximate probability density function. Accordingly, the average time complexity for constructing a classifier with the proposed kernel density estimator is $O(n \log n)$. In [9], the effects of applying the proposed kernel density estimator in a bioinformatics application is addressed.

In the following part of this paper, Section II presents the novel kernel density estimator proposed in this article. Section III reports the experiments conducted to verify the theorems presented in Section II. Finally, concluding remarks are presented in section IV.

## II. The Proposed Kernel Density Estimator

In this section, we will first elaborate the mathematical basis of the novel kernel density estimator proposed in this article. In particular, we will show that the pointwise mean squared error (MSE) of the basic form of the proposed kernel density estimator converges at $O(n^{-2/3})$, regardless of the dimension of the vector space, where $n$ is the number of instances in the training dataset. Then, we will discuss how the proposed kernel density estimator can be exploited in data classification applications.

Since we can always conduct a translation operation with the coordinate system, without loss of generality, we assume in the following discussion that it is the pointwise MSE at the origin of the coordinate system that is of concern. Let $f_X(x_1, x_2, ..., x_m)$ denote the probability density function of the distribution of concern in an $m$-dimensional vector space. Assume that $f_X(x_1, x_2, ..., x_m)$ is analytic and $f_X(x_1, x_2, ..., x_m) < \infty$ for all $(x_1, x_2, ..., x_m) \in \mathbf{R}^m$. Let $Z$ be the random variable that maps a sampling instance $s_i$ taken from the distribution governed by $f_X$ to $\|\mathbf{s}_i\|^m$, where $\|\mathbf{s}_i\|$ is the distance between the origin and $s_i$. Accordingly, we have the distribution function $F_Z(z)$ of $Z$ equal to

$$\iint \cdots \int_{x_1^2 + x_2^2 + \ldots + x_m^2 \leq z^{2/m}} f_X(x_1, x_2, ..., x_m) dx_1 dx_2 \ldots dx_m$$

for $z \geq 0$ and $F_Z(z) = 0$ for $z < 0$.

**Theorem 1:** Let $f_Z(z) = \lim_{\substack{\varepsilon \to 0 \\ \varepsilon > 0}} \dfrac{F_Z(z+\varepsilon) - F_Z(z)}{\varepsilon}$ for $z \geq 0$. Then, we have

$f_Z(0) = \dfrac{\pi^{m/2}}{\Gamma(m/2+1)} f_X(\mathbf{0})$, where $\Gamma(\cdot)$ is the gamma function [10].

**Proof:**
Since $F_Z(0) = 0$, we have

$$\lim_{\substack{\varepsilon \to 0 \\ \varepsilon > 0}} \frac{F_Z(\varepsilon) - F_Z(0)}{\varepsilon}$$

$$= \lim_{\substack{\varepsilon \to 0 \\ \varepsilon > 0}} \frac{\iint \cdots \int_{x_1^2 + x_2^2 + \ldots + x_m^2 \leq \varepsilon^{2/m}} f_X(x_1, x_2, ..., x_m) dx_1 dx_2 \ldots dx_m}{\varepsilon}$$

By the Taylor expansion,
$$f_X(x_1, x_2, ..., x_m)$$
$$= f_X(\mathbf{0}) + \frac{\partial f_X(\mathbf{0})}{\partial x_1} x_1 + \ldots + \frac{\partial f_X(\mathbf{0})}{\partial x_m} x_m + \text{high - order term s.}$$

Furthermore, in region where $x_1^2 + x_2^2 + ... + x_m^2 \leq \varepsilon^{2/m}$, we have $x_1 \to 0$, $x_2 \to 0$, ..., $x_m \to 0$ as $\varepsilon \to 0$. Therefore,

$$f_Z(0) = \lim_{\substack{\varepsilon \to 0 \\ \varepsilon > 0}} f_X(\mathbf{0}) \cdot \frac{(\sqrt[m]{\varepsilon})^m \pi^{m/2}}{\Gamma(m/2+1)} \cdot \frac{1}{\varepsilon}$$

$$= \frac{\pi^{m/2}}{\Gamma(m/2+1)} \cdot f_X(\mathbf{0}),$$

where $\frac{(\sqrt[m]{\varepsilon})^m \pi^{m/2}}{\Gamma(m/2+1)}$ is the volume of a sphere in an $m$-dimensional vector space with radius $= \sqrt[m]{\varepsilon}$.

□

Theorem 1 implies that we can obtain an estimate of $f_X(\mathbf{0})$ by first obtaining an estimate of $f_Z(0)$. Since $f_Z$ is a univariate probability density function, if we employ a fixed kernel density estimator [4] to estimate $f_Z(0)$, then, as Theorem 2 shows, we can obtain an estimator of $f_X(\mathbf{0})$ with the pointwise MSE converging at $O(n^{-2/3})$.

**Theorem 2:** Let $\{s_1, s_2, ..., s_n\}$ be a set of sampling instances randomly and independently taken from the distribution governed by $f_X$ in the $m$-dimensional vector space. Assume that $f_Z(z)$ is analytic in $[0, \infty)$ with all orders of the right-sided derivatives at 0. Then, with $\sigma = \lambda \cdot n^{-1/3}$ and $\lambda$ being a positive real number,

$$\hat{f}_X(\mathbf{0}) = \sum_{i=1}^{n} \frac{1}{n} \cdot \frac{\sqrt{2} \cdot \Gamma(m/2+1)}{\pi^{(m+1)/2} \sigma} \cdot \exp\left(-\frac{\|s_i\|^{2m}}{2\sigma^2}\right)$$

is an estimator of $f_X(\mathbf{0})$ with the pointwise MSE converging at $O(n^{-2/3})$.

**Proof:**

Let $z_i = \|s_i\|^m$ and $\hat{f}_Z(0) = \sum_{i=1}^{n} \frac{1}{n} \cdot \frac{\sqrt{2}}{\sqrt{\pi}\sigma} \exp\left(-\frac{z_i^2}{2\sigma^2}\right)$ with $\sigma = \lambda \cdot n^{-1/3}$. We have

$$MSE[\hat{f}_Z(0)] = (E[\hat{f}_Z(0)] - f_Z(0))^2 + Var[\hat{f}_Z(0)]$$

and

$$E[\hat{f}_Z(0)] = \sum_{i=1}^{n} \int_0^\infty \frac{1}{n} \cdot \frac{\sqrt{2}}{\sqrt{\pi}\sigma} \exp\left(-\frac{z^2}{2\sigma^2}\right) f_Z(z) dz.$$

As $n \to \infty$, we have $\sigma \to 0$ and

$$O\left(E[\hat{f}_Z(0)] - f_Z(0)\right)$$
$$= O\left(\left[2 \cdot \int_0^\infty \frac{1}{\sqrt{2\pi}\sigma} \exp\left(-\frac{z^2}{2\sigma^2}\right)[f_Z(0) + f'_Z(0^+)z]dz\right] - f_Z(0)\right)$$
$$= O\left(\sqrt{\frac{2}{\pi}} \cdot f'_Z(0^+) \cdot \sigma\right),$$

where

$$f'_Z(0^+) = \lim_{\substack{\varepsilon \to 0 \\ \varepsilon > 0}} \frac{f_Z(\varepsilon) - f_Z(0)}{\varepsilon}.$$

Let

$$\hat{f}_{1/n}(0) = \frac{1}{n} \cdot \frac{\sqrt{2}}{\sqrt{\pi}\sigma} \exp\left(-\frac{z_1^2}{2\sigma^2}\right).$$

We have

$$E[\hat{f}_{1/n}^2(0)] = \int_0^\infty \frac{1}{n^2} \cdot \frac{2}{\pi\sigma^2} \exp\left(-\frac{z^2}{\sigma^2}\right) f_Z(z)dz.$$

Due to $\sigma \to 0$ as $n \to \infty$,

$$O\left(E[\hat{f}_{1/n}^2(0)]\right) = O\left(\int_0^\infty \frac{1}{n^2} \cdot \frac{2}{\pi\sigma^2} \exp\left(-\frac{z^2}{\sigma^2}\right) f_Z(0)dz\right)$$
$$= O\left(\frac{f_Z(0)}{n^2 \sigma \sqrt{\pi}}\right).$$

Therefore, as $n \to \infty$,

$$O\left(Var[\hat{f}_{1/n}(0)]\right) = O\left(E[\hat{f}_{1/n}^2(0)] - (E[\hat{f}_{1/n}(0)])^2\right)$$
$$= O\left(\frac{f_Z(0)}{n^2 \sigma \sqrt{\pi}} - \frac{1}{n^2}(E[\hat{f}_Z(0)])^2\right).$$

Since $\sigma = \lambda \cdot n^{-1/3}$, as $n \to \infty$, we have

$$O\left(Var[\hat{f}_{1/n}(0)]\right) = O\left(n^{-5/3}\right).$$

Furthermore, since $s_1, s_2, \ldots, s_n$ are taken randomly and independently,

$$Var[\hat{f}_Z(0)] = n \cdot Var[\hat{f}_{1/n}(0)].$$

Therefore, as $n \to \infty$,

$$O\left(MSE[\hat{f}_Z(0)]\right)$$
$$= O\left((E[\hat{f}_Z(0)] - f_Z(0))^2 + Var[\hat{f}_Z(0)]\right)$$
$$= O(n^{-2/3}).$$

Let

$$\hat{f}_X(\mathbf{0}) = \sum_{i=1}^n \frac{1}{n} \cdot \frac{\sqrt{2} \cdot \Gamma(m/2+1)}{\pi^{(m+1)/2}\sigma} \cdot \exp\left(-\frac{\|s_i\|^{2m}}{2\sigma^2}\right)$$

with $\sigma = \lambda \cdot n^{-1/3}$. Then, we have

$$MSE[\hat{f}_X(\mathbf{0})] = E[(\hat{f}_X(\mathbf{0}) - f_X(\mathbf{0}))^2]$$

$$= E\left[\left(\frac{\Gamma(m/2+1)}{\pi^{m/2}}\right)^2 \cdot (\hat{f}_Z(0) - f_Z(0))^2\right]$$

$$= \left(\frac{\Gamma(m/2+1)}{\pi^{m/2}}\right)^2 \cdot MSE[\hat{f}_Z(0)]$$

Since $MSE[\hat{f}_Z(0)]$ converges at $O(n^{-2/3})$ with $\sigma = \lambda \cdot n^{-1/3}$, $MSE[\hat{f}_X(\mathbf{0})]$ converges at $O(n^{-2/3})$ as well.

$\square$

**Theorem 3:** Let $\{s_1, s_2 \ldots, s_n\}$ be a set of sampling instances randomly and independently taken from the distribution governed by $f_X$ in the $m$-dimensional vector space. Let

$$\hat{f}_X(\mathbf{0}) = \sum_{i=1}^{n} \frac{1}{n} \cdot \frac{\sqrt{2} \cdot \Gamma(m/2+1)}{\pi^{(m+1)/2}\sigma} \cdot \exp\left(-\frac{\|s_i\|^{2m}}{2\sigma^2}\right).$$

We have

$$\int_{-\infty}^{\infty}\int_{-\infty}^{\infty}\ldots\int_{-\infty}^{\infty} \hat{f}_X(\mathbf{0}) dx_1 dx_2 \ldots dx_m = 1.$$

**Proof:**

In order to prove

$$\int_{-\infty}^{\infty}\int_{-\infty}^{\infty}\ldots\int_{-\infty}^{\infty} \hat{f}_X(\mathbf{0}) dx_1 dx_2 \ldots dx_m = 1,$$

we only need to show that the kernel function employed satisfies

$$\int_{-\infty}^{\infty}\int_{-\infty}^{\infty}\ldots\int_{-\infty}^{\infty} \frac{\sqrt{2} \cdot \Gamma(m/2+1)}{\pi^{(m+1)/2}\sigma} \cdot \exp\left(-\frac{(x_1^2 + x_2^2 + \ldots + x_m^2)^m}{2\sigma^2}\right) dx_1 dx_2 \ldots dx_m = 1.$$

We have

$$\int_{-\infty}^{\infty}\int_{-\infty}^{\infty}\ldots\int_{-\infty}^{\infty} \exp\left(-\frac{(x_1^2 + x_2^2 + \ldots + x_m^2)^m}{2\sigma^2}\right) dx_1 dx_2 \ldots dx_m$$

$$= \int_0^{\infty} \frac{2 \cdot \pi^{m/2} \cdot r^{m-1}}{\Gamma(m/2)} \exp\left(-\frac{r^{2m}}{2\sigma^2}\right) dr,$$

where $\frac{2 \cdot \pi^{m/2} \cdot r^{m-1}}{\Gamma(m/2)}$ is the surface area of a sphere with radius $r$ in an $m$-dimensional vector space.

Let $t = r^m$.

Then, $\frac{dt}{dr} = m \cdot r^{m-1} = m \cdot \frac{t}{r}$.

Accordingly,

$$\int_0^{\infty} \frac{2 \cdot \pi^{m/2} \cdot r^{m-1}}{\Gamma(m/2)} \exp\left(-\frac{r^{2m}}{2\sigma^2}\right) dr$$

$$= \int_0^\infty \frac{2 \cdot \pi^{m/2}}{\Gamma(m/2)} \cdot \frac{1}{m} \cdot \exp\left(-\frac{t^2}{2\sigma^2}\right) dt = \frac{\pi^{m/2}}{\Gamma(m/2+1)} \int_0^\infty \exp\left(-\frac{t^2}{2\sigma^2}\right) dt$$

$$= \frac{\pi^{m/2}}{\Gamma(m/2+1)} \cdot \frac{\sqrt{\pi} \cdot \sigma}{\sqrt{2}}.$$

Therefore,

$$\int_{-\infty}^\infty \int_{-\infty}^\infty \cdots \int_{-\infty}^\infty \frac{\sqrt{2} \cdot \Gamma(m/2+1)}{\pi^{(m+1)/2} \sigma} \cdot \exp\left(-\frac{(x_1^2 + x_2^2 + \ldots + x_m^2)^m}{2\sigma^2}\right) dx_1 dx_2 \ldots dx_m = 1.$$

□

The validity of Theorem 2 stems on the assumption that $f_Z(z)$ is analytic in [0, ∞) with all orders of the right-sided derivatives at 0. One may wonder how strict this condition is. In this respect, the following illustration should provide us with some insights. Assume that $f_X(x_1, x_2, \ldots, x_m)$ is a constant function in the proximity of 0. Then, $f_Z(z)$ is analytic in [0, $\varepsilon$], where $\varepsilon$ is a small positive real number, with the right-sided derivatives at 0.

The estimator presented in Theorem 2 forms the basis of the novel kernel density estimator proposed in this article. Since both Theorem 1 and Theorem 2 address only the pointwise MSE, for real applications we have incorporated the basic idea of variable kernel density estimator [2, 4] to generalize the estimator presented in Theorem 2 and obtain the so-called super-radius based kernel density estimator (SRKDE) shown in the following:

$$\hat{f}_X^*(\mathbf{v}) = \frac{\Gamma(m/2+1)}{\pi^{m/2}} \cdot \sum_{i=1}^n \frac{1}{n} \cdot \frac{\sqrt{2}}{\sqrt{\pi} \sigma_i} \exp\left(-\frac{\|\mathbf{s}_i - \mathbf{v}\|^{2m}}{2\sigma_i^2}\right),$$

where

1. $\sigma_i = \beta \frac{[R_k(\mathbf{s}_i)]^m}{k}$;
2. $\beta$ is the smoothing parameter with order $O(n^{2/3})$;
3. $R_k(\mathbf{s}_i)$ is the distance from $\mathbf{s}_i$ to its $k$-th nearest neighbor;
4. $k$ is a parameter to be set.

The proposed kernel density estimator is so named because random variable $Z$ maps a sampling instance $\mathbf{s}_i$ taken from the distribution governed by $f_X$ to $\|\mathbf{s}_i\|^m$ and $\|\mathbf{s}_i\|^m$ is referred to as the super-radius of $\mathbf{s}_i$ in this article. For data classification applications, we will construct one SRKDE to approximate the distribution of one class of training instances in the vector space. Then, a query instance located at $\mathbf{v}$ is predicted to belong to the class that gives the maximum value among the likelihood functions defined in the following:

$$L_j(\mathbf{v}) = \frac{|S_j| \cdot \hat{f}_j^*(\mathbf{v})}{\sum_h |S_h| \cdot \hat{f}_h^*(\mathbf{v})},$$

where $|S_j|$ is the number of class-*j* training instances and $\hat{f}_j^*(v)$ is the SRKDE corresponding to class-*j* training instances. In our current implementation, aiming to improve the execution time of the classifier, we include only a limited number, denoted by *k'*, of the nearest class-*j* training instances of *v* in computing $\hat{f}_j^*(v)$.

The basic idea of the proposed kernel density estimator can be exploited to obtain a kernel-based approximate function. Let $\{s_1, s_2, ..., s_n\}$ be a set of sampling instances randomly and independently taken from the space of function *f* with a uniform sampling density $\rho$. Then, with $\sigma = \lambda \cdot n^{-1/3}$ and $\lambda$ being a positive real number, the pointwise MSE of the following kernel-based approximate function in an *m*-dimensional vector space

$$\hat{f}(0) = \sum_{i=1}^{n} \frac{f(s_i)}{\rho} \cdot \frac{\sqrt{2} \cdot \Gamma(m/2+1)}{\pi^{(m+1)/2} \cdot \sigma} \exp\left(-\frac{\|s_i\|^{2m}}{2\sigma^2}\right)$$

converges at $O(n^{-2/3})$ regardless of the value of *m*.

As mentioned earlier, one main distinctive property of the kernel density estimation based approach is that the average time taken to construct a classifier is in the order of $O(n \log n)$, where *n* is the total number of training instances. This argument is based on the assumption that the kd-tree structure [11] is employed in the implementation. For detailed analysis of the time complexity, please refer to the discussion presented in [7], which provides the detailed analysis for a similar kernel density estimator. Concerning the execution time for making prediction with *n'* query instances, it is shown in [7] that the average time complexity is $O(n' \log n)$.

## III. Experimental Results

This section reports the experiments conducted to verify the theorems presented in the previous section.

Table 1: The observed MSEs with the estimator presented in Theorem 2 at 5 different points and the respective convergence rates.

| n <br> Points | 20000 | 80000 | 320000 | 1280000 | c in <br> log*MSE* = c log*n* + δ |
|---|---|---|---|---|---|
| (0,0,0,0) | 3.23E-05 | 1.43E-05 | 5.98E-06 | 2.18E-06 | -0.643 |
| (0.05,0,0,0) | 3.45E-05 | 1.48E-05 | 5.12E-06 | 2.50E-06 | -0.644 |
| (0,1,0,0) | 2.23E-05 | 8.1E-06 | 3.55E-06 | 1.38E-06 | -0.661 |
| (0,0.1,0,0) | 4.08E-05 | 1.23E-05 | 6.22E-06 | 2.47E-06 | -0.656 |
| (0.05,0.05,0,0) | 3.57E-05 | 1.43E-05 | 5.70E-06 | 2.41E-06 | -0.649 |

Each dataset used in the experiment contained sampling instances randomly taken from the distribution defined by the following probability density function in the 4-dimensional vector space:

$$\frac{1}{(\sqrt{2\pi})^4} \cdot \left\{ \frac{11}{20} \exp\left(-\frac{[(x_1-0.1)^2 + x_2^2 + x_3^2 + x_4^2]}{2}\right) + \frac{9}{21} \exp\left(-\frac{[(x_1+0.1)^2 + x_2^2 + x_3^2 + x_4^2]}{2}\right) \right\}$$

Then, the estimator presented in Theorem 2 with $\sigma$ set to $0.005 \cdot \left(\frac{n}{10000}\right)^{-1/3}$ was employed to obtain the estimates at the 5 points listed in Table 1 based on the randomly generated dataset. For the numbers reported in Table 1, the same experimental procedure was repeated 500 times with 500 independently generated datasets and the observed MSE at each point was computed as follows

$$\frac{1}{500} \sum_{i=1}^{500} \left(\hat{f}_i(v) - f(v)\right)^2,$$

where $\hat{f}_i(v)$ is the estimate of $f(v)$ obtained with the dataset generated in the $i$-th run of the experiment. The experimental results reported in Table 1 confirm that the pointwise MSE of the estimator presented in Theorem 2 converges at $O(n^{-2/3})$.

## IV. Conclusion

This article proposes the super radius based kernel density estimator (SRKDE) and reports the experiment conducted to verify the theorems presented in this article. The major distinction of the SRKDE is that the pointwise MSE of its basic form converges at $O(n^{-2/3})$, where $n$ is the number of instances in the training dataset, regardless of the dimension of the vector space, if the probability density function at the point of interest meets certain conditions. Since the average time complexity for constructing a SRKDE based classifier is $O(n\log n)$, it is conceivable that the SRKDE based approach can cope well with the contemporary applications that involve a large and ever-growing database and delivers ever-improving prediction accuracy as the database continues to grow. On the other hand, because the theorems associated with the SRKDE are all derived with the asymptotical approach, these theorems may not hold well when the number of sampling instances is not sufficiently large. In such cases, the SRKDE based classifier may deliver inferior prediction accuracy in comparison with the state-of-art SVM (Support Vector Machine). In summary, the kernel density estimation based approach and the SVM have their respective advantages and disadvantages.

## Contributions of Authors

YJO initiated this study, proposed the SRKDE, and established its mathematical foundation. DTHC, YYO, HGH, CPW and CYC jointly implemented the software and designed the experiments reported in this article.

## Acknowledgement

The authors greatly appreciate the comments and suggestions provided by Prof. Henry Horng-Shing Lu of National Chiao-Tung University.  This research has been supported by the National Science Council of R.O.C. under the contracts NSC 95-3114-P-002-005-Y and NSC 96-2627-B-002-003.